\documentclass{article}
\usepackage{ijcai20}

\usepackage{times}

\usepackage{soul}
\usepackage{url}
\usepackage[hidelinks]{hyperref}
\usepackage[utf8]{inputenc}
\usepackage[small]{caption}
\usepackage[justification=centering]{caption}
\usepackage{graphicx}
\usepackage{amsmath}
\usepackage{booktabs}
\title{Deep Conversational Recommender System:\\ A New Frontier for Goal-Oriented Dialogue Systems}

\author{
Dai Hoang Tran$^1$\footnote{Contact Author}\and
Quan Z. Sheng$^1$\and
Wei Emma Zhang$^2$\and
Salma Abdalla Hamad$^1$\and\\
Munazza Zaib$^1$\and
Nguyen H. Tran$^3$\and
Lina Yao$^4$\And
Nguyen Lu Dang Khoa$^5$\\
\affiliations
$^1$Department of Computing, Macquarie University\\
$^2$School of Computer Science, The University of Adelaide\\
$^3$School of Computer Science, The University of Sydney\\
$^4$School of Computer Science and Engineering, The University of New South Wales\\
$^5$Data61, CSIRO\\
\emails
\{dai-hoang.tran, michael.sheng\}@mq.edu.au, wei.e.zhang@adelaide.edu.au, \\\{salma-abdalla-ibrahim-mah.h, munazza-zaib.ghori\}@hdr.mq.edu.au,\\ nguyen.tran@sydney.edu.au, lina.yao@unsw.edu.au, khoa.nguyen@data61.csiro.au
}

\begin{document}

\maketitle

\begin{abstract}

In recent years, the emerging topics of recommender systems that take advantage of natural language processing techniques have attracted much attention, and one of their applications is the Conversational Recommender System (CRS). Unlike traditional recommender systems with content-based and collaborative filtering approaches, CRS learns and models user's preferences through interactive dialogue conversations. In this work, we provide a summarization on the recent evolution of CRS, where deep learning approaches are applied to CRS and have produced fruitful results. We first analyze the research problems and present key challenges in the development of Deep Conversational Recommender Systems (DCRS), then present the current state of the field taken from the most recent researches, including the most common deep learning models that benefit DCRS. Finally, we discuss future directions of this vibrant area.

\end{abstract}

\section{Introduction}
In recent years, natural language processing techniques have advanced by leaps and bounds, and we are witnessing the booming of conversational user interfaces via virtual agents from big companies such as Microsoft Cortana, Amazon Alexa, Apple Siri, and Google Assistant. These agents can perform multiple tasks via voice or text commands. Even though their capabilities are still primitive, the array of actions they can carry out are impressive. In another area, recommender systems have emerged as a separated research field in the 90's \cite{Next_Generation_of_Recommender_Systems}, which are now acting as the core functionality of some of the largest online services in the world such as YouTube, Netflix and Amazon Book \cite{Sequential_Recommender_Systems,YouTube_Recommendations}. Initially, recommender system techniques were mainly based on content-based and collaborative filtering approaches \cite{Deep_Learning_Based_Recommender_System}. However, the need of conversational systems that can provide good suggestions to users is essential to many online e-commerce services, thus establishing the ground for the development of conversational recommender systems (CRS) \cite{Towards_Conversational_Recommender_Systems}. This can be seen as a natural extension of these conversational virtual agents, and researchers are trying to make this new application a reality. Over the years, we have seen several approaches to develop a CRS, from using contextual bandit to machine learning methods \cite{Towards_Conversational_Recommender_Systems,Making_Personalized_Recommendation}. Nevertheless, we are seeing a recent trend in the field of CRS, where deep learning approaches are being used to provide end-to-end solutions for CRS, and these systems are considered as Deep Conversational Recommender Systems (DCRS) \cite{Towards_Deep_Conversational}. To better understand the DCRS, in this paper, we provide an overview of what CRS and DCRS are, what are the challenges pertinent to the development of DCRS, what are the current state-of-the-art deep models for DCRS, and discuss its future research directions. All the information is collected from papers in the top conferences in the past five years. To the best of our knowledge, our work is the first one that tries to summarize and understand the DCRS in details.
 
\section{Deep Conversational Recommender Systems} \label{sec_overview}

\subsection{Backgrounds}
\begin{figure}[t]
	\centering
	\includegraphics[width=1\linewidth]{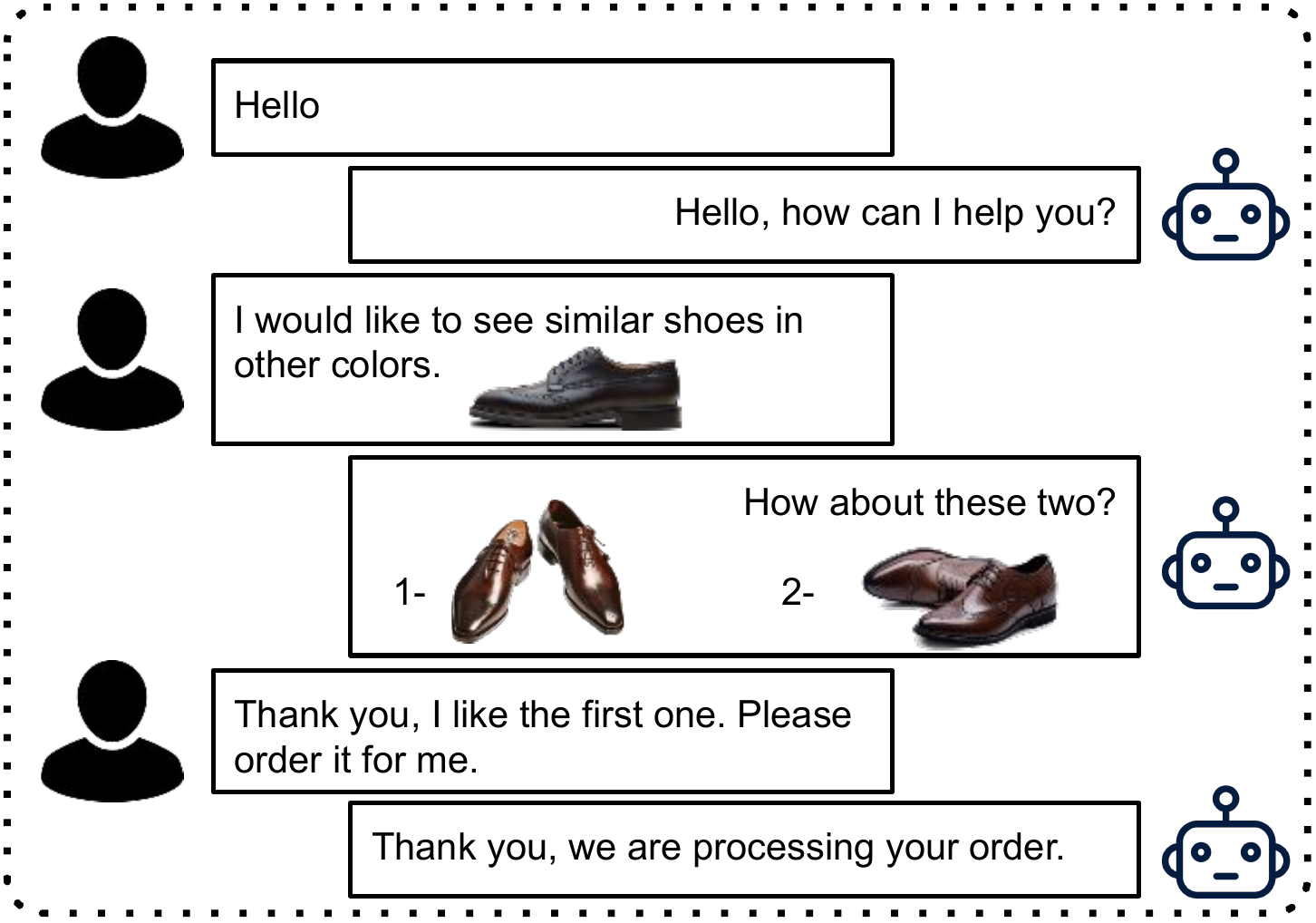}
	\caption{An example conversation between a user and an agent in a Conversational Recommender System in e-Commerce domain.}
	\label{fig_crs_interface}

\end{figure}

In daily life activities, human's most natural interactive actions are communicating with others via conversations. We converse about our work, we gossip about other people's relationship, and we recommend our friends about things we like. When it comes to recommendations, from seeking advice from our friends for good movies to watch, to looking for enjoyable holiday destinations from travel agents, we can express our preferences and quickly get recommendations from others through just a few exchanges of simple conversations. Here we use the term \textit{users} to denote the information seekers, and the term \textit{agents} to denote the information providers.
From the perspective of online businesses, due to the natural and personal characteristics of direct communication via conversations,
a large amount of modern services provide call or chat systems to deal with customer support. However, human resources are limited and costly, thus, the need of intelligent agents who are able to converse with users and give satisfactory recommendations is essential to them. On the other hand, from the user's perspective, the ability to freely express one's preferences and retrieve tailored suggestions from the agents give the users a strong sense of satisfaction and confidence in the choices they make. These conversational systems that provide tailored suggestions to the users and can carry out intelligent conversations are called CRS. An example of a CRS dialogue session is illustrated in Figure \ref{fig_crs_interface}, which depicts a scenario where the agent is able to learn and provide matching shoes recommendation to a satisfactory user.

\begin{figure}[t]
	\centering
	\includegraphics[width=1\linewidth]{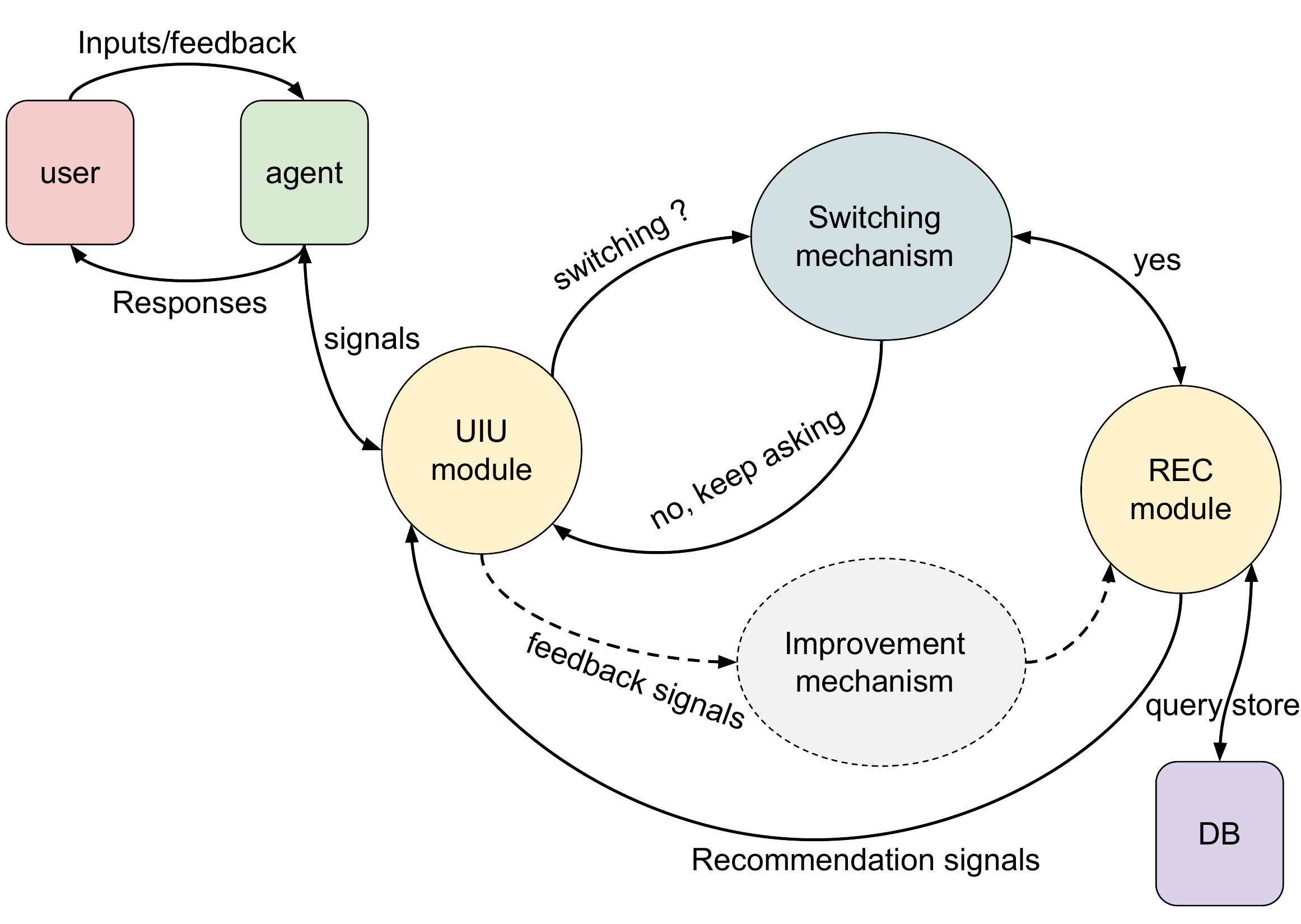}
	\caption{The main components of a DCRS. The dash line shows optional component of a DCRS.}
	\label{fig_dcrs_general_architecture}

\end{figure}

In a typical CRS, there exists three components, a \textit{User Intention Understanding} (UIU) module, a \textit{Recommendation} (REC) module, and a \textit{Switching Mechanism} (SWM). User's inputs and agent's responses are handled by the UIU module. Since the most common form of user's inputs and agent's responses is text, most of the models have trained UIU module that can process and generate natural language data. Yet we are seeing an increasing number of works on UIU that can handle \textit{multimodal} inputs, such as both text and image inputs, which will be discussed in later sections. UIU produces dialogue states which will be consumed and decided by the SWM to either keep asking users more questions for clarification, or pass the dialogue state to the REC module to generate recommendations. Notice that in some cases, the SWM receives both signals from UIU and REC modules to make a decision \cite{Conversational_Recommender_System}. Optionally, certain researches also proposed \textit{Improvement Mechanism} to improve the system's recommendations based on user's feedback \cite{Estimation_Action_Reflection}. To achieve these complex goals, the UIU and the REC modules are usually trained by traditional machine learning approaches \cite{Towards_Conversational_Recommender_Systems}. In recent years, the tremendous success of deep learning methods in a variety of tasks have encouraged the development of Deep CRS, whereas the UIU and the REC's training methods are replaced by deep learning approaches, leading to performance improvement in these systems. Thus we define DCRS as a CRS that has at least one of its modules using deep learning approach to develop the system. Figure \ref{fig_dcrs_general_architecture} shows the main components of a DCRS.

\subsection{Definitions and Formalization} \label{sec_definition}
DCRS is primarily a goal-oriented dialogue system which also incorporates the element of chit-chat and question answering dialogue systems \cite{Towards_Deep_Conversational}. Figure \ref{fig_crs_interface} illustrates such an example. This natural mode of interaction between users and agents presents considerable challenges when designing such a system. Concretely, a DCRS takes \textit{users} $u \in U$ inputs about certain facets of possible items via an \textit{utterance}, then the \textit{agent} $a$ responds with either a new question to learn more about current user's preferences, or recommended items. At each \textit{turn} $t$, the user $u$ provides utterance $s_{ut}$, and the agent gives responded utterance at next turn $s_{a(t+1)}$. A dialogue \textit{session} is a collection of these utterances $S_t=\{s_{u1}, s_{a2},\dots,s_{ut},s_{a(t+1)}\}$ until the user stops responding or terminates the session. In each session, there can be multiple \textit{rounds} of conversation, each round ends when the agent provides recommended items. If the user initiates new request after that, a new round starts. A \textit{multi-round} DCRS is a system where after each round, the agent can improve its recommendation outputs based on the learned user's preferences from previous round \cite{Estimation_Action_Reflection}. Some DCRS accept user's \textit{feedback} at the end of a round. A user's \textit{feedback} is usually a positive or negative confirmation utterance after the agent's recommendation output. 

At each turn $t$, the utterance set $S_t$ is fed into the UIU module to produce a dialogue state embedding with $d$-dimension ${z_t} \in R^{d}$, usually via a Recurrent Neural Network (RNN) based encoder. The SWM module uses the ${z_t}$ embedding to output an action score with $n$-dimension $p \in R^n$, which depends on the policy of the system, the $p$ score will be classified into an action that is either (i) \textit{more-question action} or (ii) \textit{give-recommendation action}. For the first action of \textit{more-question}, the UIU module generates agent's responded question to ask the user for more specific information, usually an attribute-related question. For the second action of \textit{make-recommendation}, the REC module then queries its database to search for appropriate items based on dialogue state ${z_t}$, and returns the recommendation signal ${r_t}$ to UIU module. The UIU then generates agent's responded utterance containing the recommendation, and this marks the end of one round of conversation. The work of \cite{Q_R} follows this flow. In certain cases, to provide more accurate responses, at turn $t$, both UIU and REC modules receive the $S_t$ signal and produce $z_t$ and $r_t$ signals respectively. The signals are then combined and used by SWM to decide the policy action \cite{System_Ask_User_Respond}. The cycle repeats. 

\section{DCRS Research Problems and Challenges} \label{sec_dcrs_charac}
Developing a DCRS is challenging due to its inherit nature of having two different tasks in one system, which are (i) understanding user intention, and (ii) giving relevant recommendations. The whole system needs to understand user intention, user's preferences and optionally, user's feedback through natural language, which is a challenging problem by itself. Apart from that, the DCRS also needs to know how to query its database to find relevant and personalized items based on the user's inputs. Table \ref{table_summary} summarizes key characteristics of prominent literature works. In this section, we outline the current research problems and challenges of DCRS. 

\begin{table*}[t]
	\small
	\centering
	
	\begin{tabular}{|l|l|l|l|l|l|l|l|l|}
		\hline
		\textbf{\textbf{\textbf{DCRS}}} & \textbf{\textbf{\textbf{Dataset}}} & \textbf{\textbf{\textbf{\begin{tabular}[c]{@{}l@{}}Question \\ Space\end{tabular}}}} & \textbf{\textbf{FR$^1$}} & \textbf{\textbf{\textbf{SWM}}} & \textbf{\textbf{P$^2$}} & \textbf{\textbf{\textbf{\begin{tabular}[c]{@{}l@{}}Multi-\\ modal\end{tabular}}}} & \textbf{\textbf{\textbf{\begin{tabular}[c]{@{}l@{}}Multi-\\ round\end{tabular}}}} & \textbf{\textbf{\textbf{\begin{tabular}[c]{@{}l@{}}Neural\\ Networks\end{tabular}}}} \\ \hline
		\begin{tabular}[c]{@{}l@{}}VisualDialog (KDD) \\ \cite{A_Visual_Dialog_Augmented}\end{tabular} & synthetic & texts & no & rule-based & no & yes & yes & \begin{tabular}[c]{@{}l@{}}RNN-base,\\ CNN-based\end{tabular} \\ \hline
		\begin{tabular}[c]{@{}l@{}}KMD (ACM-MM)\\ \cite{Knowledge_aware_Multimodal_Dialogue}\end{tabular} & MMD$^*$ & texts & yes & policy network & no & yes & yes & \begin{tabular}[c]{@{}l@{}}HRED, EITree, \\ Reinforcement\end{tabular} \\ \hline
		\begin{tabular}[c]{@{}l@{}}DeepCR (NeurIPS) \\ \cite{Towards_Deep_Conversational}\end{tabular} & ReDIAL & texts & yes & pointer softmax & yes & no & yes & \begin{tabular}[c]{@{}l@{}}HRED, \\ Autoencoder\end{tabular} \\ \hline
		\begin{tabular}[c]{@{}l@{}}DCR (CoRR)\\ \cite{Deep_Conversational_Recommender_in_Travel}\end{tabular} & MultiWOZ$^{**}$ & texts & yes & pointer softmax & no & no & no & \begin{tabular}[c]{@{}l@{}}HRED,\\ GCN\end{tabular} \\ \hline
		\begin{tabular}[c]{@{}l@{}}SAUR (CIKM) \\ \cite{System_Ask_User_Respond}\end{tabular} & synthetic & attributes & no & rule-based & no & no & yes & \begin{tabular}[c]{@{}l@{}}Memory \\ Network\end{tabular} \\ \hline
		\begin{tabular}[c]{@{}l@{}}CRM (SIGIR) \\ \cite{Conversational_Recommender_System}\end{tabular} & synthetic & attributes & no & policy network & yes & no & no & \begin{tabular}[c]{@{}l@{}}LSTM,\\ Reinforcement\end{tabular} \\ \hline
		\begin{tabular}[c]{@{}l@{}}EAR (WSDM) \\ \cite{Estimation_Action_Reflection}\end{tabular} & synthetic & attributes & no & policy network & yes & no & yes & Reinforcement \\ \hline
		\begin{tabular}[c]{@{}l@{}}Q\&R (KDD) \\ \cite{Q_R}\end{tabular} & Youtube & attributes & no & rule-based & yes & no & no & RNN-based \\ \hline
		\begin{tabular}[c]{@{}l@{}}CEI (AI*IA)\\ \cite{Converse_Et_Impera}\end{tabular} & synthetic & texts & no & policy network & no & no & yes & \begin{tabular}[c]{@{}l@{}}RNN-based, \\ Reinforcement\end{tabular} \\ \hline
	\end{tabular}
	
	\caption[caption]{Summarization of prominent DCRS in the literature with their characteristics.
		\\FR$^1$ represents the Fluent Response of agent, P$^2$ denotes Personalization.
		\\MMD$^*$ \cite{Towards_Building_Large_Scale_Multimodal}, MultiWOZ$^{**}$ \cite{MultiWOZ}}
	\label{table_summary}

\end{table*}

\subsection{Understanding User's Intention}
During a dialogue session in a DCRS, a user keeps expressing her intention via natural language, it can be as simple as a greeting to initiate a conversation, or to explain her preferences for items, or to give feedback to the agent. It is critical for the agent to know what is the current user's intention for the system to act. However, text understanding and comprehension is a complicated ongoing research in the field of deep learning \cite{Towards_AI_Complete_Question_Answering}. 
Due to this issue, \textit{understanding user's intention} is one of the key challenges in developing DCRS.

We observe that currently, there are two groups of approaches being used to understand a user intention. The first group, which is typically used by DCRS that are trained on synthetic dataset, is attributes extraction from user's utterances. Assuming that after each question asked by an agent, the user utterance will contain $V=\{v_1, v_2,\dots, v_n\}$ attributes from a fix set of attributes $A$. Then the set $V$ will be fed to the UIU module. This approach simplifies the encoding aspect of the user's utterance, as can be seen in these works \cite{Q_R,Estimation_Action_Reflection}. The second group tries to encode the whole dialogue utterances via RNN-based neural network to extract a dialogue state $z \in R^d$ to feed to the UIU module, and this approach is better at generating fluent responses by an agent \cite{Towards_Deep_Conversational,Knowledge_aware_Multimodal_Dialogue}.

\subsection{Providing Personalized Recommendations}
People might assume that a DCRS always provides personalized recommendations, but that is not always the case. The first priority of a DCRS is to provide relevant recommendations, which matches user's preferences. However, certain DCRS are designed to act more like a search-and-filter engines, that only consider user's preferences in the current dialogue session. Thus, two users with the same preferences might receive the same recommendations. To overcome this issue of lacking personalized recommendations in the system, the DCRS needs to know the user's features (age, gender, etc.), to remember the user's feedback as well as her past preferences, and finally incorporate those signals with dialogue state to generate personalized recommendations. Hence, two users with the same preferences still receive different recommendations due to their history differences. It is important to have a \textit{personalized DCRS} for better user engagement.

In the current literature, we have seen a mix of these solutions. Some works solely act as a search or filter engine without taking user's attributes into account \cite{A_Visual_Dialog_Augmented,System_Ask_User_Respond}, while others do provide personalized recommendations using user's attributes in their systems \cite{Conversational_Recommender_System,Estimation_Action_Reflection}.

\subsection{Developing Suitable Switching Mechanism} \label{sec_swm}
It is imperative for a DCRS to know when to ask questions to learn more about user's preferences, or when to give responses with recommendation results. Because if this mechanism behaves wrongly, it will lead to lengthy conversation or inappropriate recommendation turn, which results in user's dissatisfaction. Not only that, SWM is also the connection between major modules in a DCRS as illustrated in Figure \ref{fig_dcrs_general_architecture}. Hence, it is a key challenge for developing a high performance DCRS, that is to provide an \textit{accurate SWM}. 

So far, we have observed three groups of approaches for SWM, namely \textit{rule-based, pointer softmax probability, and reinforcement policy score}. For rule-based SWM, there is no specific methodology. A rule can be just a simple constraint, such as for each turn, always providing a recommendation \cite{Q_R}, or a designed choice such as the confident score of the top-k recommended items over a threshold \cite{System_Ask_User_Respond}. Pointer softmax probability is originated from \cite{Pointing_the_Unknown_Words}, where SWM uses the Gated Recurrent Unit (GRU) cell to decode hidden state of the current dialogue context, and decides whether to generate response tokens with or without recommended item via a softmax probability score, which leads to natural and fluent agent's generative responses as seen in the works of \cite{Towards_Deep_Conversational,Deep_Conversational_Recommender_in_Travel}. The third approach is reinforcement policy score, where at each turn in a dialogue session, SWM generates an action vector based on embedding signals from UIU and REC modules. This action vector then is fed into a reinforcement policy network, and outputs a softmax class score to reflect which action to take. Henceforth, the policy network is trained to maximize the action reward based on the labels of the dataset. This SWM approach is chosen by the works that focus on generating mechanical responses as seen in \cite{Conversational_Recommender_System,Estimation_Action_Reflection,Converse_Et_Impera}.

\subsection{Handling of Multimodal Inputs/Outputs}
The user's input to a DCRS is often text. However, modern conversational systems allow multiple input types such as text, image or audio files (e.g., Facebook Messenger). In this regards, it is also natural for DCRS to be able to process \textit{multimodal} inputs aside from textual input, and the agents need to be able to provide multimodal outputs as well. It is a real challenge for DCRS since it introduces another complexity into the UIU module, where the agent has to understand the semantic meaning of other non-textual input types. For instance, if the user expresses she wants to find similar dresses from one clothing image, the agent needs to understand various features of that dress to find similar ones to recommend. Different input types require specialized methods to extract their semantic meanings, which can also require additional training of sub-module from the whole model.

As such, currently, there are only a handful of researches that address such challenge and a few notable works are \cite{Knowledge_aware_Multimodal_Dialogue,A_Visual_Dialog_Augmented}. In \cite{Knowledge_aware_Multimodal_Dialogue}, the author's DCRS can incorporate image semantic meaning through an EI-tree neural network as well as external domain knowledge. The work of \cite{A_Visual_Dialog_Augmented}, on the other hand, provides unique user's response via item clicking, then all of these signals including user's utterances and user's requested images, are passed to an Augmented Cascading Bandit module to provide agent's responses. We expect to see more research works to tackle multimodal data in the future.

\subsection{Training Multi-Task Models}
As mentioned in Section \ref{sec_overview}, DCRS contains multiple modules to handle different tasks, notably the UIU and REC modules. Even though the main objective is to provide relevant recommended items via agent's responses, it does include sub-goals such as the agent utterance generation and the finding of top-k recommended items based on user's utterances. Henceforth, how to \textit{incorporate multiple objective functions of different tasks for an end-to-end model training} is one of the key challenges in developing DCRS.

From our investigation during the survey, the majority of researches contain element of optimizing multi-task loss function in the form of $L_{\{t1,\dots,tn\}} = L_{t1} + \dots + L_{tn}$. Usually, the multitask loss function $L_{\{t1,\dots,tn\}}$ is placed at SWM, to optimize the agent's best action to be taken at each turn of the dialogue session. Whereas the other loss functions $L_t$ is optimized separately from their respective modules \cite{Converse_Et_Impera,Knowledge_aware_Multimodal_Dialogue,Estimation_Action_Reflection}.

\subsection{Training on Limited Fluent Dialogue Datasets} \label{sec_dataset_challenge}
The first and foremost challenge for any deep learning problem is having a large and accurate dataset, and in the field of DCRS, it is a major issue. Through our survey of recent works, DCRS is lacking in the fluent dialogue datasets. A fluent dialogue dataset contains  natural and fluent human-to-human conversation. However, only a handful of them are available in limited quantity such as ReDIAL \cite{Towards_Deep_Conversational} and MultiWOZ \cite{MultiWOZ} datasets. As a deep learning model needs lots of training data, synthetic DCRS dataset is the preferable choice for several research works because it is simple to bootstrap and generate, as shown in Table \ref{table_summary}. Due to this lacking of natural fluent dialogue conversation datasets, it is unavoidable that solutions based on synthetic datasets will get exposed to biased or non-fluent/mechanical responses. As such, \textit{having a big dataset of fluent conversion in DCRS} is a key challenge when developing a DCRS.

This has allowed us to observe an interesting fact, the researches that utilize fluent dialogue datasets tend to focus on generating \textit{seamless} responses in \textit{multi-round} setting, where the agent's response can include zero or more recommended items, while the user can keep carrying out the conversation in a natural way \cite{Towards_Deep_Conversational,Deep_Conversational_Recommender_in_Travel}. On the other hand, the researches that use synthetic datasets for validation tend to focus on generating mechanical yet relevant questions to exploit user's preferences. Thus if the user's answer does not match with the training templates, the agent will not understand and may ask the same question again \cite{Estimation_Action_Reflection,System_Ask_User_Respond}. Therefore, having a big dialogue datasets with natural conversation is important for developing a fluent and accurate DCRS. We are seeing several works trying to address this issue by using machine learning to generate more realistic dialogues such as \cite{Chat_More,An_Automatic_Procedure}.

\section{DCRS Deep Learning Models}
\begin{figure}[t]
	\centering
	\includegraphics[width=1\linewidth]{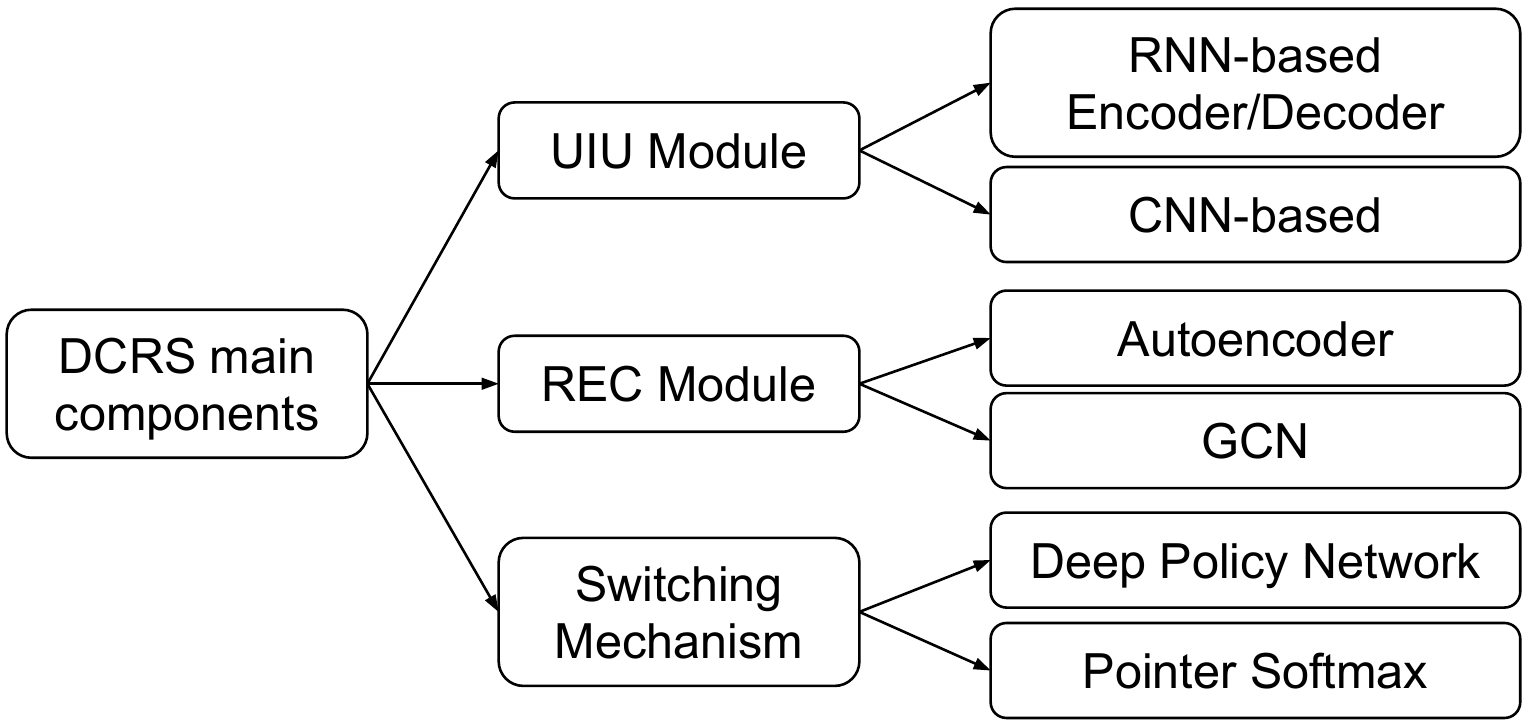}
	\caption{Deep learning approaches for each DCRS component.}
	\label{fig_dcrs_technical}

\end{figure}

In this section, we provide an in-depth look at the deep learning approaches for developing DCRS. When we are categorizing the technical aspects into buckets, we realize that it is not easy to divide them based on the whole architecture of the system, since a DCRS contains multiple components, and each of them can have its own deep learning model. Thus, we make the categorization based on deep learning models applied for three main components of DCRS, which are the UIU module, REC module and SWM, as shown in Figure \ref{fig_dcrs_technical}.
Next, we elaborate the details of how these deep learning models act in their respective modules.

\subsection{UIU Deep Learning Models}
A required task of a DCRS is to understand user's intention via her inputs. Due to its conversational nature, most of UIU models are deep learning models for understanding textual natural language inputs. However, few notable works also try to tackle multimodal data such as both text and image inputs \cite{A_Visual_Dialog_Augmented,Knowledge_aware_Multimodal_Dialogue}. We will outline the most popular deep learning approaches to handle user's utterances in DCRS in the following.

\subsubsection{RNN-based Models}
Due to the popularity of RNN-based models for dealing with natural languages, the majority of the UIU models are RNN-based models in the DCRS of our survey. The most popular one is the hierarchical recurrent encoder decoder (HRED) model as seen in the works of \cite{Towards_Deep_Conversational,Deep_Conversational_Recommender_in_Travel,Knowledge_aware_Multimodal_Dialogue}. Other RNN variants are the  memory network with attention weight in the work of \cite{System_Ask_User_Respond,Knowledge_aware_Multimodal_Dialogue}, and GRU encoder in \cite{Converse_Et_Impera}. The basic operation of these models is to encode the current dialogue session of $S_t$ of turn $t$ into a dialogue state vector $z_t$, which will be further processed by other modules of the DCRS. 

\subsubsection{CNN-based Models}
Multimodal DCRS uses the CNN-based deep learning model to extract image input features for encoding dialogue state. The extracted image features are usually concatenated with utterance features to be processed in further pipeline. In \cite{A_Visual_Dialog_Augmented}, the authors used a pre-trained ResNet CNN model to extract image features, while the work of \cite{Knowledge_aware_Multimodal_Dialogue} has its CNN model to extract semantic meaning of image, called EI-tree.

\subsection{REC Deep Learning Models}
Many REC modules of different DCRS do not use deep learning models. Instead, they opt for Matrix Factorization approach, due to its capability to take advantage of all users and item's attributes \cite{Conversational_Recommender_System,Estimation_Action_Reflection}. Nevertheless, we observe a few variants of deep learning models used for making recommendations in DCRS.

\subsubsection{Autoencoder-based Models}
Autoencoder has been used in the research of general recommender systems to overcome the cold-start problem, and a notable work using this model is Autorec \cite{AutoRec}. Based on this approach, the work of \cite{Towards_Deep_Conversational} train their REC module using a deep de-noising autoencoder network for predicting user's ratings that have not been observed in the training set.

\subsubsection{GCN-based Models}
Graph Convolutional Neural Network (GCN) deep learning models allow us to solve problems based on graph-structured data, such as social network or recommendation item relationship \cite{GraphSAGE}.
The principle of GCN network is that it takes into account both node's attributes as well as node's neighbourhood attributes using graph convolution operation, thus allows the model to learn better local representation of each node, and achieves state-of-the-art performance in graph-structured data tasks \cite{Web_Scale_Recommender_Systems}. Base on this, the work of \cite{Deep_Conversational_Recommender_in_Travel} builds a GCN recommendation model for a travel DCRS application, which constructs the graph-structured data to connect hotels, restaurants, and other travel facilities to provide additional services to customers that go well together.

\subsection{Switching Mechanism Deep Learning Models}
SWM plays an important role in keeping the DCRS behaving correctly at each user's utterance input as explained in Section \ref{sec_swm}. Certain researches applied simple rule-based methods such as providing recommendation after each turn \cite{Q_R}, or returning recommendations when the first top-k items ranking probability reaches a certain threshold \cite{System_Ask_User_Respond}. But researchers are using more sophisticated approaches to train the SWM, and the deep learning models they often use are deep policy network and pointer softmax probability.

\subsubsection{Deep Policy Network}
Deep policy network is a straight-forward usage of reinforcement learning to maximize the reward of a SWM action based on current dialogue state. The action space is usually either \textit{more-question action} or \textit{give-recommendation action} as mentioned in Section \ref{sec_definition}. For each dialogue turn during the training phase, a reward score is given to the SWM to make correct action choice, and negative reward is given otherwise. Both the works of \cite{Conversational_Recommender_System,Estimation_Action_Reflection} use a two-layer feed-forward neural network to optimize the network parameters, while \cite{Knowledge_aware_Multimodal_Dialogue} use BLUE score as the reward signal.

\subsubsection{Pointer Softmax Probability}This technique is based on the work of \cite{Pointing_the_Unknown_Words}. The principle of this method is that during the inference process, if a certain rare condition is met, the neural network can point to other data points that it knows how to handle, instead of processing the current rare condition. By utilizing this approach in the SWM, a DCRS can  generate fluent agent's responses with recommendation. As proved in the work of \cite{Towards_Deep_Conversational}, at step $m$ in the dialogue $D$, the UIU module keeps generating sentence tokens while asking the SWM if it should point to a movie name or not, based on the current dialogue state. If the SWM decides to point to a movie name, the UIU will generate recommended movie names and add to the generative response of the agent. The work of \cite{Deep_Conversational_Recommender_in_Travel} also uses similar principle.

\section{Open Research Directions}
The prosperity of deep learning in advancing natural language processing and recommendation tasks have brought more development to the field of DCRS. Especially in the past five years, the number of works on DCRS have increased tremendously. Some challenges still remain unsolved and need more research efforts. We have identify further research directions, and discuss in the following sections.

\subsection{Synthetic Fluent Dialogue Datasets}
As one of the challenges in developing a DCRS, we need better datasets for training the DCRS deep learning models. However, given the high cost and complication of making a real human-to-human dialogue dataset, it is more feasible to create high quality synthetic dialogue datasets. With the advancement of deep learning in handling natural language tasks such as language generation, text comprehension, and question-answering \cite{Read_Verify,BERT}, we strongly believe that researchers can use deep learning techniques to make better synthetic training datasets for the development of DCRS.

\subsection{Incorporating External Domain Knowledge}
A particular work of \cite{Knowledge_aware_Multimodal_Dialogue} proposes a unique approach by using external domain knowledge to improve their DCRS. The authors embed external knowledge domain into a memory network, then generate agent's responses based on extracted domain knowledge and the current dialogue context. Given the lacking of good training datasets in DCRS, being able to incorporate external domain knowledge from free knowledge bases such as DBPedia\footnote{dbpedia.org} or NELL\footnote{rtw.ml.cmu.edu} can definitely bring benefits to the DCRS area. Henceforth, one main direction is to develop methods for better integrating external domain knowledge into the DCRS architecture to increase system's performance.

\subsection{Improvement from User's Feedback}
In the current literature of DCRS, we find that not many works consider user's feedback to improve system performance. The work of \cite{Estimation_Action_Reflection} addresses this concern via a reflection phase, where negative feedback is collected and stored for future re-training of the system. It is a simple and effective approach. Another approach is learning user's feedback via intent taxonomy \cite{User_Feedback_Intents}. 
Therefore, future researchers can develop online methods to improve DCRS directly from user's feedback for better system performance.

\subsection{Unified Evaluation Metrics} 
A noticeable observation from our study is the discrepancy of measurement metrics when evaluating a DCRS. This occurs due to the inherent multitask nature of the DCRS. To the best of our knowledge, there is no \textit{single metric to evaluate the DCRS as a whole}. Therefore, researchers rely on both goal-oriented dialogue and recommendation measurement metrics to evaluate their works. Some metrics are used more than the others such as BLUE score to evaluate the fluency of agent's generative responses \cite{Knowledge_aware_Multimodal_Dialogue,Converse_Et_Impera}, while several others use the success-rate metric to measure their recommendation's effectiveness \cite{Conversational_Recommender_System,Estimation_Action_Reflection}. We hope to see more unified measurement metrics for DCRS evaluation.

\section{Conclusion}
Conversational recommender systems are a practical application domain for modern online services, and deep conversational recommender systems are the next evolution of CRS. In recent years, we have seen a rising effort of researches that aim to improve this new exciting field. By taking a detailed look at the current state of the field in different angles, summarizing their characteristics, problems, challenges and proposing future research directions, we hope that our survey provides useful information and elicits excitement for more researchers to contribute to this vibrant research area.

\bibliographystyle{named}
\bibliography{ijcai_7pages}

\end{document}